\documentclass[letterpaper, 10 pt, conference]{ieeeconf}  
\IEEEoverridecommandlockouts                              

\usepackage{amsmath,epsfig}
\usepackage{graphicx} 
\usepackage{epstopdf}
\usepackage{xcolor}
\usepackage{tabularx}
\usepackage{multirow}
\usepackage{threeparttable}
\usepackage{subfigure}
\PassOptionsToPackage{bookmarks=false}{hyperref}

\usepackage{amsmath} 
\usepackage{arydshln}

\usepackage{cite}

\usepackage[colorlinks=true,
            linkcolor=red,
            urlcolor=blue,
            citecolor=gray]{hyperref}

\definecolor{dgreen}{rgb}{0, 0.6, 0}

\title{\LARGE \bf
BORM: Bayesian Object Relation Model for Indoor Scene Recognition  
}
\author{~~~~~{Liguang Zhou$^{1,2}$, Jun Cen$^{2,3}$, Xingchao Wang$^{1,2}$, Zhenglong Sun$^{1,2}$, Tin Lun Lam$^{1,2,\dagger}$, Yangsheng Xu$^{1,2}$}
\thanks{$^{\dagger}$ Corresponding Author: Tin Lun Lam (tllam@cuhk.edu.cn)}
\thanks{This work was supported in part by the funding AC01202101025 and 2019-INT007 from the Shenzhen Institute of Artificial Intelligence and Robotics for Society.}
\thanks{\textsuperscript{1} School of Science and Engineering, The Chinese University of Hong Kong, Shenzhen}
\thanks{\textsuperscript{2}  Shenzhen Institute of Artificial Intelligence and Robotics for Society, The Chinese University of Hong Kong, Shenzhen.}
\thanks{\textsuperscript{3} Hong Kong University of Science and Technology}
}

\begin{document}
\maketitle
\thispagestyle{empty}
\pagestyle{empty}


\begin{abstract}
Scene recognition is a fundamental task in robotic perception. For human beings, scene recognition is reasonable because they have abundant object knowledge of the real world. The idea of transferring prior object knowledge from humans to scene recognition is significant but still less exploited. 
In this paper, we propose to utilize meaningful object representations for indoor scene representation. First, we utilize an improved object model (IOM) as a baseline that enriches the object knowledge by introducing a scene parsing algorithm pretrained on the ADE20K dataset with rich object categories related to the indoor scene. To analyze the object co-occurrences and pairwise object relations, we formulate the IOM from a Bayesian perspective as the Bayesian object relation model (BORM). Meanwhile, we incorporate the proposed BORM with the PlacesCNN model as the combined Bayesian object relation model (CBORM) for scene recognition and significantly outperforms the state-of-the-art methods on the reduced Places365 dataset, and SUN RGB-D dataset without retraining, showing the excellent generalization ability of the proposed method. Code can be found at \href{https://github.com/hszhoushen/borm}{\textcolor{magenta}{https://github.com/FreeformRobotics/BORM}}.
\end{abstract}


\section{Introduction}
\label{sec:intro}

Scene understanding is a fundamental cognitive ability for robots to conduct tasks in an unknown environment. To carry out the task commanded by controllers, the most fundamental problem is recognizing the environment robots are traveling in, such as the living room, bedroom, or kitchen.

Recognizing scene about the current environment is quite essential for robots to make more intelligent behaviors. Fortunately, scene recognition has been an important research area among computer vision and robotics for decades. There have been many scene recognition algorithms proposed that using visual information for scene understanding. Liu et al. \cite{liu2009scene}\cite{Liu2013} propose a color and geometric features based adaptive descriptor for scene recognition. Also, indoor scenes can be characterized by global spatial features \cite{quattoni2009recognizing}. However, these methods are based on the traditional lower-level features, which are not accurate enough, and not interpretable at the semantic level.

\begin{figure}[tbp]
	\centering
	\includegraphics[width=0.48\textwidth]{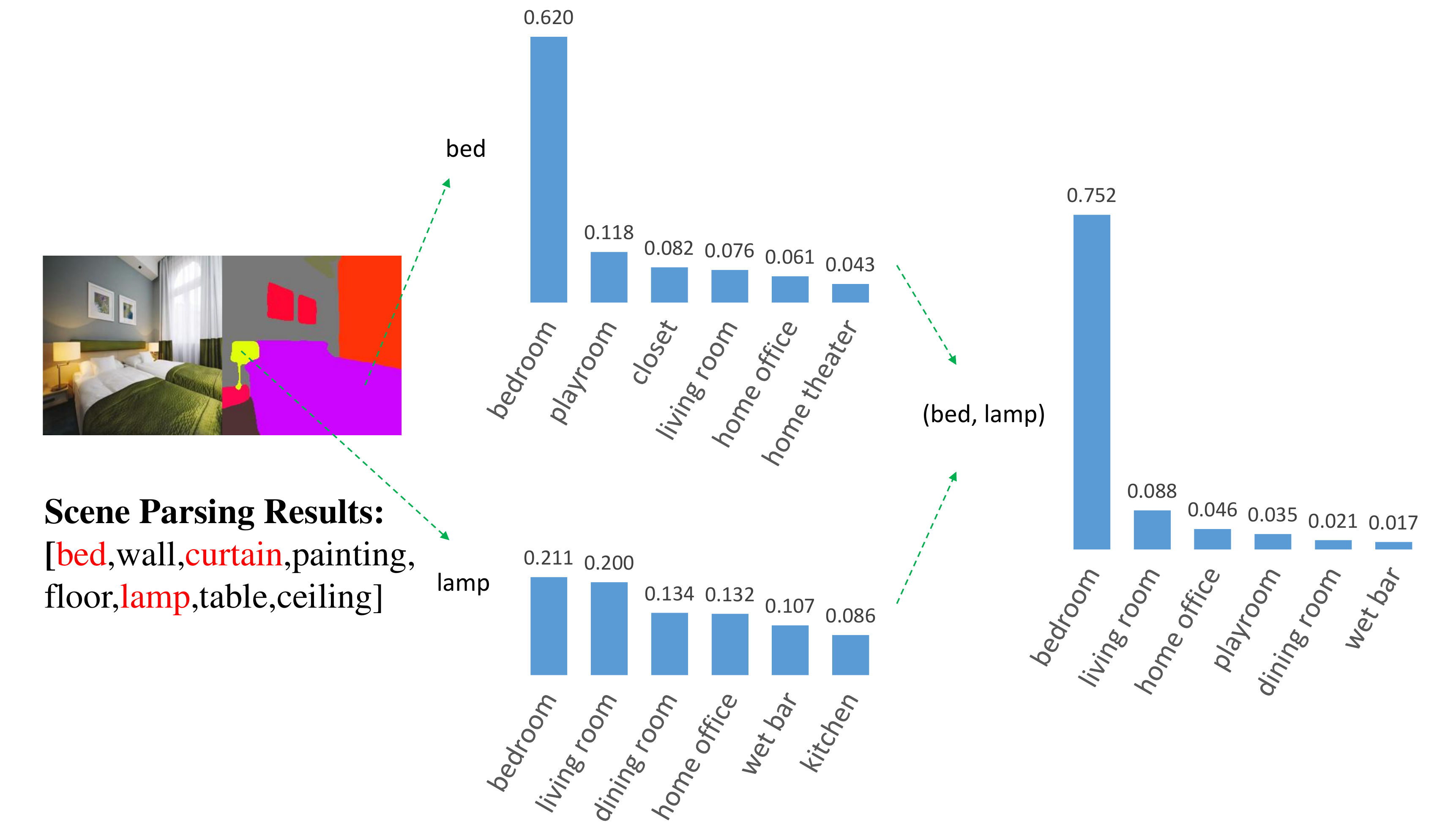}
	\caption{The left part of the figure shows the input image and the corresponding scene parsing result. We have conducted a Bayesian probabilistic analysis using the BORM on the Places365-14 dataset. The middle part shows the conditional probability of $P(scene|object)$, and the top 6 scene given the object ``bed" and ``lamp" are presented, respectively. The right part shows the joint conditional probability of $P(scene | (object, object))$, and the top 6 scene given the object pair $(bed, lamp)$. As observed from the figure, the proposed BORM suppresses the probability of scene-common object pairs while highlights the probability of the scene-specific object pairs.}
	\label{fig:joint_prob}
\end{figure}

Recently, many deep learning based methods have demonstrated impressive performance over various computer vision tasks. For example, ResNet has been shown a convincing ability in image classification tasks for years \cite{He2016}. However, the classification of the image using ResNet is like using a black box to interpret the image, which is not really in an interpretable manner.  Chen et al. \cite{chen2019scene} consider the word-embedding model to reformulate the scene understanding problem in a more semantic meaningful perspective. The output of ResNet module, object detection module, and scene parsing module have been encoded in the one global vector for scene understanding. However, the performance of the word embedding method is just slightly higher than ResNet, as reported. Besides, there are three streams used for scene recognition, which is relatively inefficient.
The utilization of semantic information for scene recognition is important. To model the descriptors probabilities, a Bayesian filtering method is proposed \cite{wu2009visual}, but the object information is not explicitly used. To use the object information, a object classifier is used to classify low-level visual features to objects \cite{espinace2010indoor}, while the relation among objects is not exploited. To utilize the object relation for scene recognition, spatial object-to-object relation is studied for RGB-D scene recognition \cite{song2019image}. Besides, a Long Short-Term Memory modeling method is proposed to investigate the object relation with ROI selection \cite{laranjeira2019modeling}. To utilize the object information in the scene, the object model \cite{pal2019deduce} is proposed as complementary semantic information of the scene combined with ResNet to better interpret the given scene. However, using only the object vector might not be discriminative enough for scene understanding \cite{zeng2019learning}. A Object-to-Scene model is proposed, where the object features and object relation are learned by object feature aggregation module and object attention module \cite{Miao2021ots}, respectively. Overall, these above-mentioned method lack the statistically analysis of object distribution in the scene. Specifically, there are some scene-common objects across various scene and scene-specific objects only appear in the particular scene. Therefore, we propose the Bayesian object relation model (BORM) for improving the accuracy of scene recognition by highlighting the scene-specific objects while preserving the scene-common objects for scene understanding \cite{cheng2018scene}. However, all of these methods neglect the probabilistic relation among object pairs, which contains an essential message for scene recognition.

In this paper, we consider the probabilistic relation between the object pairs as shown in Fig. \ref{fig:joint_prob}. We can observed that there is a high probability of the specific object pair appear in the particular scene, e.g., the bed and lamp is most likely (75.2\%) appeared in the bedroom, and can be regarded as a scene-specific object pair. 

To utilize this conditional object pair relation w.r.t the various scenes, the BORM that derives the joint conditional probability of the object pairs given the various scenes. With the BORM, the joint conditional probability of the scene-specific object pairs will be enhanced and the joint conditional probability of scene-common object pairs will be suppressed.

In summary, our main contributions of this paper are as follows:

\begin{itemize}
	\item We utilize an IOM as baseline, based on scene parsing algorithm pretrained on ADE20K dataset that contains more relevant object classes for indoor scene representation. Surprisingly, the IOM surpasses the object model over \textbf{20.5\%} accuracy on average. 
	
	\item Meanwhile, To conduct an in-depth study of object pair relations, we propose a Bayesian object relation model (BORM) that enhances the scene-specific object pair relation and suppresses the scene-common object pair relation in a Bayesian probability manner for indoor scene representation.
	
	\item We combine the proposed BORM and the PlacesCNN model as CBORM, which significantly outperforms the state-of-the-art methods on the Places365-7 and Places365-14 dataset over 2.0\% and 2.1\% accuracy, and SUN RGB-D dataset over 2.0\% without retraining the model, showing the excellent generalization ability of the proposed method.

\end{itemize}

The rest of the paper is organized as follows. Section~\ref{sec:related_work} introduces the related work of scene recognition and object knowledge for indoor scene representation. Section~\ref{sec:method} describes IOM that pretrained on ADE20K with 150 categories based on the scene parsing algorithm, and the BORM that conduct an in-depth study of object relation in a Bayesian perspective. In Section~\ref{sec:cBORM}, we discuss the PlacesCNN model for scene recognition. Moreover, we combine the proposed BORM and the PlacesCNN model as the CBORM for scene recognition.  Section~\ref{sec:exp} shows the experimental settings and results, and numerical experiments have been conducted to verify the effectiveness of our proposed method. Finally, the conclusions and future directions are pointed out in Section~\ref{sec:conclusion}.

\begin{figure*}[htb]
	\centering
	\includegraphics[width=0.9\textwidth]{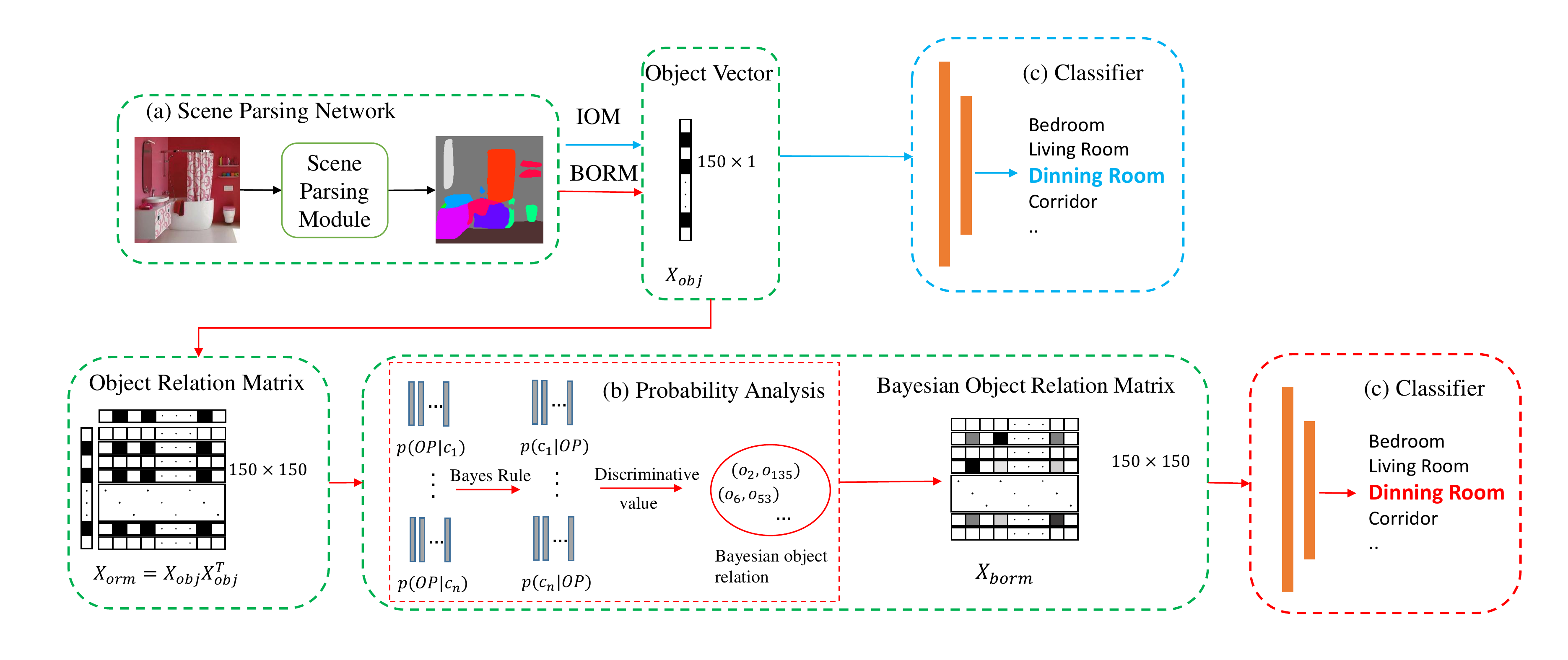}
	\caption{BORM that based on the baseline IOM is shown in the diagram, where blue and red arrow shows the flow of the IOM and BORM, respectively.}
	\label{fig:avom}
\end{figure*}

\section{Related Work}
\label{sec:related_work}
In this section, we review the research works related to our paper in two aspects: scene recognition, and object knowledge from object detection or scene parsing algorithm for indoor scene representation. We also discuss the differences and connections between these related works and our method.

\subsection{Scene Recognition}
Scene recognition has been an important research problem in robotics area for decades, which can be utilized for topological map construction and mobile robot localization \cite{liu2009scene}\cite{Liu2013}\cite{Lin2018}. Moreover, it has the potential to be applied for robot to perform efficient recognition of functional areas \cite{Ye2017}, identification of the person \cite{wang2019learning}, and execute task accordingly. The early methods, mainly focus on extraction of local features like color descriptors \cite{Liu2013}\cite{VanDeSande2009}. 
To better recognize indoor images, the combination of local and global features are utilized \cite{quattoni2009recognizing}. However, these methods only captures lower-level features of the scene while the high-level semantic structures are difficult to capture \cite{Zhu2016}. 

To utilize the high-level semantic information, some methods propose to leverage the mid-level concepts. e.g., Zhou et al. \cite{Zhou2014}\cite{zhou2017places} propose a CNN based classifier to learn deep features for scene recognition on the Places Dataset. Liao et al. \cite{Liao2016} use deep learning with a multi-task training method that incorporate both semantic segmentation and scene recognition tasks. Zhu et al. \cite{Zhu2016} propose a discriminative multi-modal feature fusion framework for scene recognition. However, these methods neglect the critical object information for scene recognition.

To incorporate object information, Li et al. \cite{Li2014} represent images by using objects appearing in them as object bank (OB) method. Brucker et al. \cite{Brucker2018} counts the co-occurrence frequencies as potentials in conditional random field for scene labeling.
In DEDUCE \cite{pal2019deduce}, one hot object vector is used as complementary information for scene recognition, where the object information is separately considered. In context based Word Embeddings \cite{chen2019scene}, there are three streams for scene recognition, one stream is the scene parsing model pretrained on ADE20K, the other is ResNet50 pretrained on reduced Places365 dataset, and a Word Vectors Module computes the content of two modules and refines the results. Song et al. \cite{Song2017} considers spatial object-to-object relations with the intermediate (object) representations. In Semantic-aware method \cite{Lopez-Cifuentes2020}, the image representation and context information are combined, where context information consists of scene objects and stuff, and their relative locations. However, all of these methods have not considered the probabilistic relation of the object pairs given the various scene.
In this work, we consider the object pair co-occurrences given various scene in a Bayesian Perspective with proposed Bayesian object relation model (BORM). To the best of our knowledge, in the scene recognition task, the object relation modeled in a probabilistic perspective is less exploited.  

\subsection{Object Knowledge for Indoor Scene Representation}
Recently, object detection is a popular research area in computer vision, thanks to the emerging of the large-scale labeled data and advanced GPUs. Many excellent algorithms have been developed for object detection like the one-stage object detector, YoloV3 \cite{redmon2018yolov3}, SSD \cite{liu2016ssd} and two-stage object detector, like Joint SSD \cite{yi2018long}, Faster R-CNN \cite{Ren2015}, Mask RCNN \cite{he2017mask}, Cascaded RCNN \cite{cai2018cascade}. The one-stage detector has a higher speed while maintain the similar performance compared with two-stage detector, therefore, the YoloV3 is adopted as the part of object model in DEDUCE \cite{pal2019deduce}. To incorporate the object knowledge for scene understanding, the object detection algorithm is first pretrained on the public available dataset. There are several mainstream dataset for object detection. PASCAL VOC \cite{Everingham2015} contains 20 categories of objects, which is very limited, and most labels are like car, bus, bicycle, airplane, and all of these are not relevant to indoor scene representation. To better represent indoor scene, DEDUCE use the YoloV3 pretrained on the MS COCO dataset \cite{lin2014microsoft}, which contains 80 object categories. However, the half of objects in MS COCO are outdoor objects like giraffe, elephant, which are not relevant to indoor scene representations. 

Scene parsing algorithm is used to segment objects and stuff in the still image, which has demonstrated surprisingly performance recently \cite{Zhou2017}. Zhou et al. \cite{Zhou2019} propose a scene parsing algorithm to detect a wide range of object and stuff in the pixel level with ADE20K Dataset, which contains 150 classes of object knowledge in pixel level. To utilize the rich object knowledge of the ADE20K Dataset, we adapt the scene parsing model pretrained on the ADE20K dataset as the improved object model (IOM), which shows a significantly improvements over the OM pretrained on the MS COCO dataset.

\section{Bayesian Object Relation Model}
\label{sec:method}

In Section \ref{sec:method}, we present the baseline model IOM that uses more object categories for object representation and the BORM that models the object pair relation, as shown in Fig.~\ref{fig:avom}. Since the object model proposed in Deduce \cite{pal2019deduce} only contains information about few categories of objects pretrained on the MS COCO dataset, which is very limited for indoor scene recognition because most of the object categories are not relevant to indoor scene representation. e.g.,  the object categories of elephant and giraffe from the super-category of animal, airplane and bus from the super-category of vehicle, traffic light and stop sign from the super-category of outdoor, and so on. In total, there are half of the objects can be categorized as outdoor objects. We believe that if the object model possesses the more rich and relevant object knowledge about the scene, the better performance of scene recognition can be obtained. Therefore, we utilize the IOM pretrained on the ADE20K dataset with rich and relevant object categories as a baseline model for indoor scene representation. To consider the probabilistic relation of object pairs, we assume some object pairs are scene-specific, and some object pairs are scene-common. To this end, we propose a novel BORM that highlights the scene-specific object pairs while suppresses the scene-common object pairs from a Bayesian probabilistic perspective.

%

\subsection{Improved Object Model (IOM)}

Different from the basic OM that only have 80 objects information of environment based on the YoloV3 pretrained on the MS COCO dataset, we present an IOM based on the scene parsing algorithm pretrained on ADE20K \cite{Zhou2017} dataset that convert the output the scene parsing algorithm to an object vector $\mathbf{X_{iom-150}}$ with 150 dimensions where the 1 means the detected objects, while 0 means the objects are not in the given image. Compared with the OM, we now have a new scene representation of 150 dimension $\mathbf{X_{iom-150}}$. The new scene representation $\mathbf{X_{iom-150}}$ will  be fed to a two-layer fully connected network for classification with the size of 32 and the number of scenes, respectively.

\begin{figure}[t]
	\centering
	\subfigure{
		\begin{minipage}{0.35\textwidth}
			\centering
			\includegraphics[width=3.6cm,height=3.6cm]{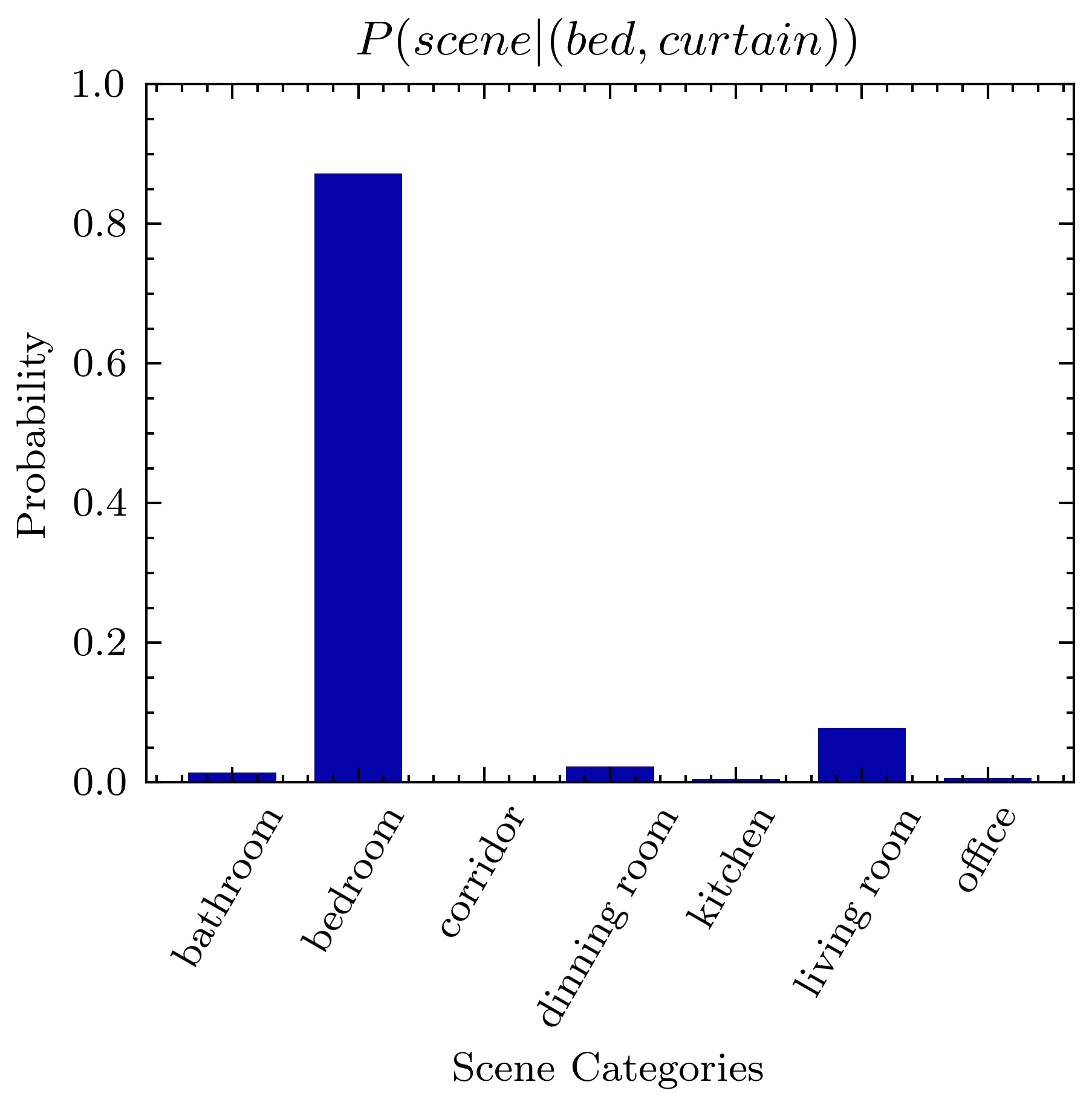}
			\label{exp:mnist:1}
	\end{minipage}}
	\subfigure{
		\begin{minipage}{0.35\textwidth}
			\centering
			\includegraphics[width=3.6cm,height=3.6cm]{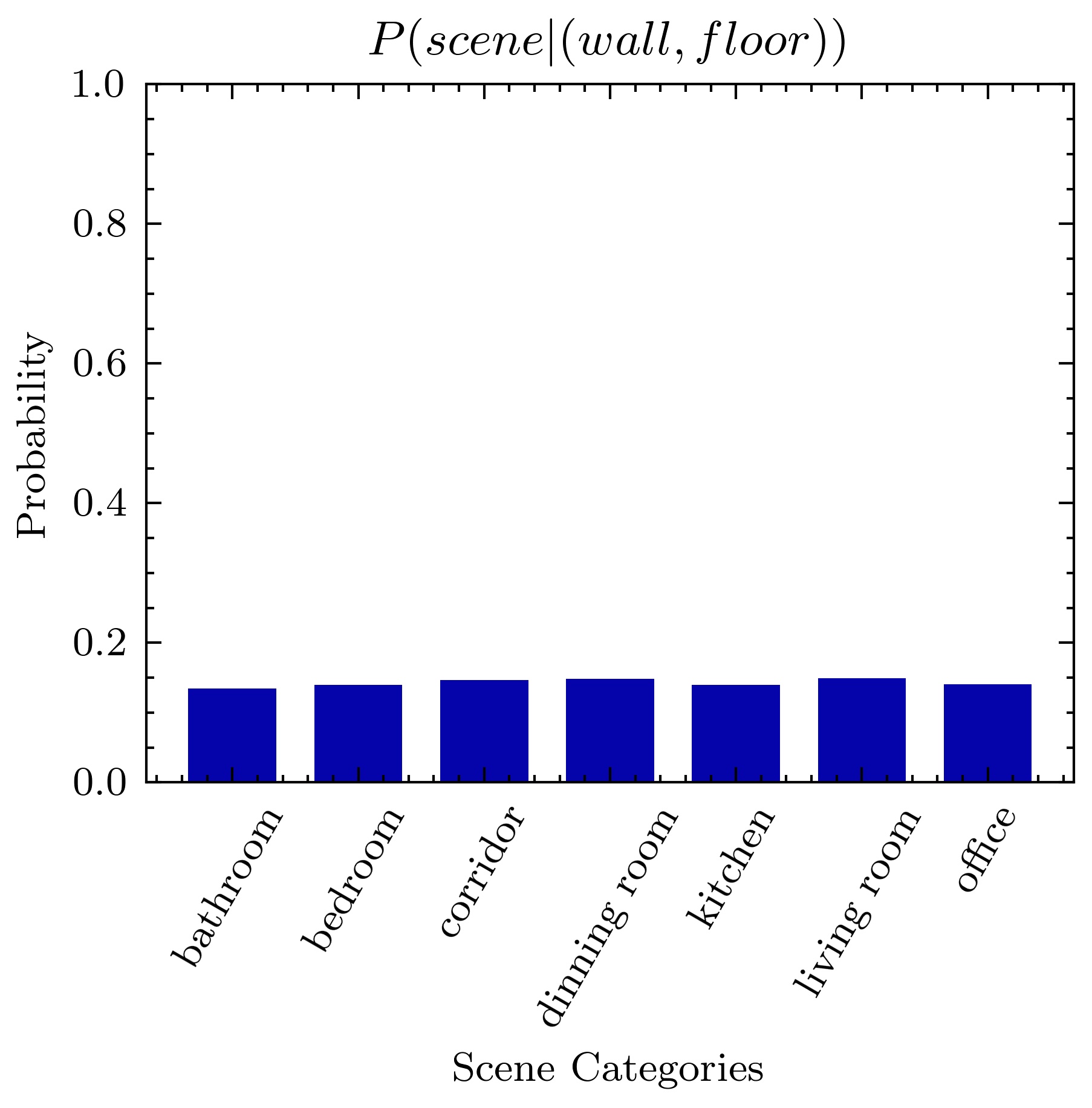}
			\label{exp:mnist:2}
	\end{minipage}}
	\caption{The BORM contains the discriminative value of object pairs. We have conducted statistical analysis on the Places365-7 dataset. The top figure shows the $p(scene|(bed, curtain))$, the probability distribution of object pair (bed, lamp) over the seven indoor scenes, and the standard deviation is 0.30, which means the $(bed, curtain)$ is a scene-specific object pair for indoor scene recognition. The bottom figure shows the $p(scene|(wall, floor))$, the probability distribution of object pair $(wall, floor)$ among seven indoor scenes,  which are almost the same. The standard deviation is 0.01, which indicates the $(wall, floor)$ is a scene-common object pair for indoor scene recognition, because this object pair appears in every scene equally.}
	\label{fig:object_pair}
\end{figure}

\subsection{Bayesian Co-occurrence Probability Analysis}
\label{sec:BORM}

\begin{figure}[htbp]
	\centering
	\includegraphics[width=0.48\textwidth]{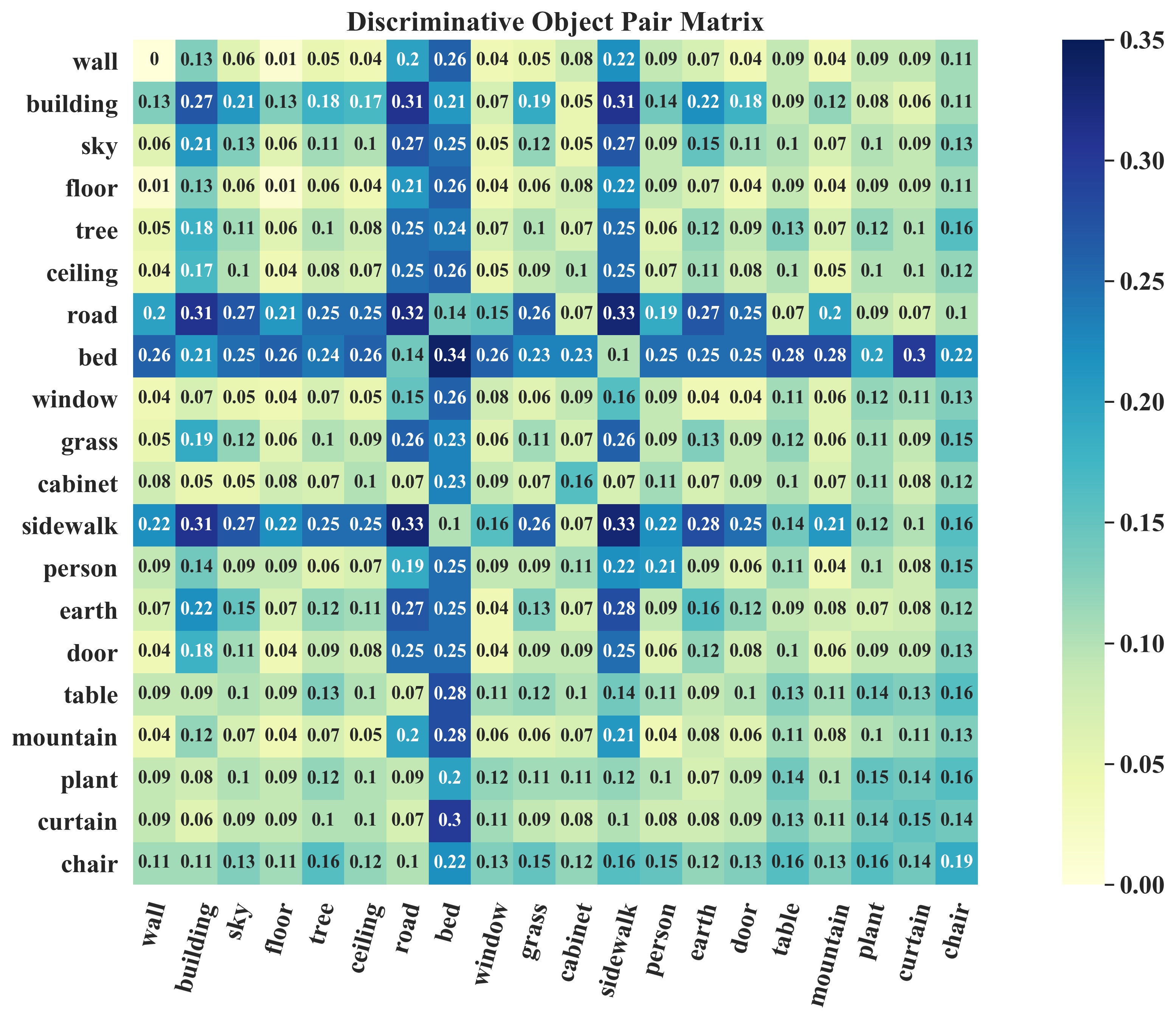}
	\caption{The Bayesian object relation matrix of first 20 objects. e.g., the  $(bed,curtain)$ is the scene-specific object pair, in which will mostly leading to the bedroom, while $(wall,floor)$ forms the scene-common object pair because it is appear in everywhere with almost same probability.}
	\label{fig:BORM}
\end{figure}

Instead of using one hot object vector as the feature representations for the scene, which lacks the probabilistic relation between different object pairs, we propose a novel BORM method that measures the probabilistic relation of object pairs in a Bayesian perspective.
The scene recognition problem is shaped by the fact that a few object pairs are scene-common, but most object pairs are scene-specific. As observed in Fig.~\ref{fig:object_pair}, the probability of scene-specific object pair $(bed, curtain)$ and scene-common object pair $(wall, floor)$ are presented.

The scene-specific object pairs indicate the object pairs that have a high probability at a particular scene and have low probability at others. e.g., the object pairs like $(bed,curtain)$ tend to be scene-specific, which means their probability $p(scene|(bed, curtain))$ can be quite distinctive among the various scene categories and has the highest probability that appears in the bedroom. Therefore, they have a large standard deviation (0.31). In contrast, scene-common object pairs distributed at various scenes with a similar probability. e.g., the common object pairs like $(wall,floor)$ frequently appear in many different scenes, which means their joint conditional probability $p(scene|(wall, floor))$ is quite similar across various scenes, i.e., they have an extremely small standard deviation (0.01) compared with those scene-specific object pairs.

\begin{figure*}[htbp]
	\centering
	\includegraphics[width=18cm]{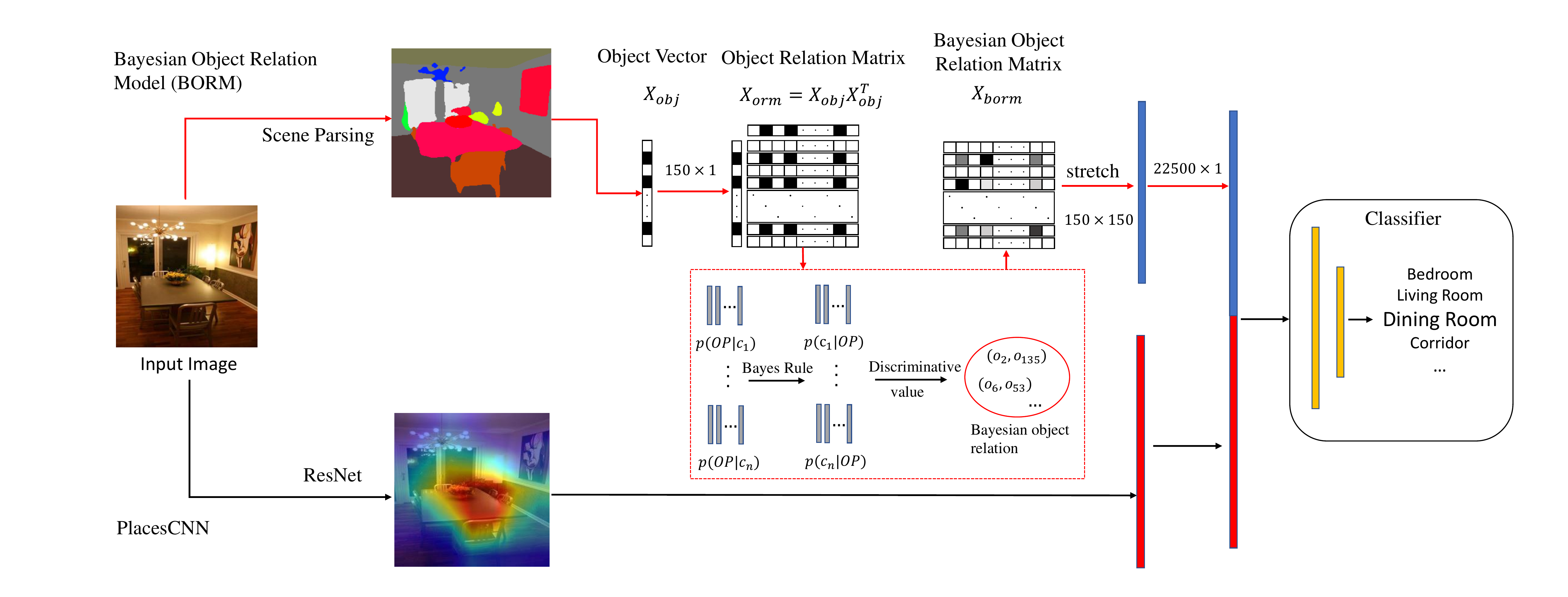}
	\caption{The proposed combined Bayesian object relation model (CBORM) contains two streams. The first stream with the red arrow is the proposed BORM that utilizes the scene parsing algorithm for scene segmentation. The second stream with the black arrow is the PlacesCNN model for feature extraction with the ResNet backbone network. The output of the BORM will first be converted to a discriminative object feature. Meanwhile, the results of the BORM and feature extraction result of PlacesCNN will be concatenated as one combined scene feature and fed to the two-layer FC network for scene classification.}
	\label{fig:combined_model}
\end{figure*}

We adopt a pipeline to estimate the posterior probability $P(c_j |o_h, o_i)$. First, we use the Places365-7 and Places365-14 dataset for learning the BORM statistically with only the training set, while testing on the SUN RGB-D dataset, where the probability distribution is regarded similar to the Places365-7 dataset. 

Given a set of images $I_{c_j}$ from a scene category $c_j$, the conditional probability of object $o_i$ appears on the scene $c_j$ is:
\begin{equation}
\begin{split}
P\left(o_{i} | c_{j}\right) = N_{o_i} / N_{I_{c_j}} \\
i \in [1,N_{objs}], j \in [1,N_{scenes}]
\end{split}
\end{equation}

where $N_{o_i}$ is the total number of i-th object $o_i$ appears in $I_{c_j}$ and $N_{I_{c_j}}$ is total number of images of $I_{c_j}$. Meanwhile, where $N_{objs}$, and $N_{scenes}$ represent the number of objects in the pretrained model and number of scene categories we have in the dataset respectively.
Assume the statistical independence of each object, we obtain the joint conditional probability $P\left(o_{h}, o_{i} | c_{j}\right)$ of $o_h$ and $o_i$ appear in the scene $c_j$:

\begin{equation}
\begin{split}
P\left(o_{h}, o_{i} | c_{j}\right) = P\left(o_{h} | c_{j}\right) P\left(o_{i} | c_{j}\right)\label{joint probability} \\
h, i \in [1,N_{objs}], j \in [1,N_{scenes}]
\end{split}
\end{equation}

The $P(c_{j}|o_{h},o_{i})$, the posterior probability of scene class $c_{j}$ given an object pair ($o_{h},o_{i}$), can be derived by the Bayes Rule and Law of Total Probability. 
\begin{equation}
\begin{split}
P\left(c_{j} | o_{h},o_{i}\right)& =\frac{P\left(o_{h},o_{i} | c_{j}\right) P\left(c_{j}\right)}{P(o_{h},o_{i})} \\
&=\frac{P\left(o_{h},o_{i} | c_{j}\right) P\left(c_{j}\right)}{\sum_{j} P\left(o_{h},o_{i} | c_{j}\right) P\left(c_{j}\right)} \\
\end{split}
\end{equation}

where the $P(c_j)$ is the probability of scene $c_j$ in the dataset, and $\sum_{j=1}^{N_{scenes}}P(c_j)=1$. We construct the posterior probability matrix $P(c_j|o,o)$ by calculating relation of each object pairs:

\begin{equation}
\scriptsize
P\left(c_{j} | o_{}, o_{}\right) =
\begin{bmatrix}
P\left( c_{j} | o_{1}, o_{1}\right) & \cdots & P\left( c_{j} | o_{1},o_{k}\right) & \cdots & P\left(c_{j} | o_{1}, o_{n}\right) \\
\vdots &  & \vdots &  & \vdots \\
P\left(c_{j}|o_{j}, o_{1}\right) & \cdots & P\left(c_{j} | o_{j},o_{k}\right) & \cdots & P\left(c_{j} | o_{j},o_{n}\right) \\
\vdots &  &  &  & \vdots \\
P\left(c_{j} | o_{n}, o_{1}\right) & \cdots & P\left(c_{j} | o_{n},o_{k}\right) & \cdots  & P\left(c_{j} | o_{n}, o_{n}\right)
\end{bmatrix}
\end{equation}



First, the posterior probability $P(c_j|o_h, o_i)$ is calculated. To calculate the discriminative value of the object pairs, the standard deviation is applied to posterior probabilities among scene categories, denoted as $std(P(c_j|o_h, o_i))$. The discriminative value of the posterior probabilities among scene categories is defined as $dis(o_h, o_i)$:


\begin{equation}
\operatorname{dis}\left(o_{h}, o_{i} \right)= std _{c \in 1, \ldots, N_{scenes}}(P(c_j|o_h, o_i))
\end{equation}

The $dis(o_h, o_i)$ is the discriminative value for measuring the discriminalibility of the object pair to the scene categories. Instead of using object vector as the feature representations for the scene, which dismiss the relationship between different objects, we first construct an object relation matrix $\mathbf{X_{orm}}=\mathbf{X_{iom}} \mathbf{X_{iom}^\mathrm{T}}$ for representing the given scene. The value of $\mathbf{X_{orm}}$ will be replaced by the discriminative value $dis(o_h, o_i)$ at the same position. Therefore, the Bayesian object relation matrix is obtained by element-wise matrix multiplication, $\mathbf{X_{borm}}=\mathbf{X_{orm}}*\mathbf{dis(o_h, o_i)}$, with the size of 150x150. After that, the matrix will be stretched to a one dimensional vector with the size of 22500x1, and will be fed into the three-layer fully connected network with the size of 8192, 2048, and number of scene classes.

As shown in Fig. \ref{fig:BORM}, the discriminative matrix of the first 20 object pairs are displayed. The $(bed,curtain)$ is a scene-specific object pair that mostly appear in the bedroom, and they form a scene-specific object pair with a discriminative value of 0.30. The $(wall, floor)$ is a common object pair appear everywhere and have an extremely small discriminative value of 0.01, which means they form a scene-common object pair. 

\section{CBORM Model}
\label{sec:cBORM}

\subsection{PlacesCNN Model}
In order to obtain the scene representation, we use the PlacesCNN model \cite{zhou2017places} with the base architecture ResNet \cite{He2016} as a backbone network, which is pretrained on the ImageNet \cite{deng2009imagenet}  dataset. There are two versions of ResNet according to the settings in DEDUCE \cite{pal2019deduce} and Word2Vec \cite{chen2019scene}. ResNet18 is used for Places365-7 and SUN RGB-D dataset, while ResNet50 is used for Places365-14 dataset. Specifically, we preserve the output of ResNet18 with 512 dimensions or ResNet50 with 2048 dimensions, namely $F_{scene}$, the feature of the PlacesCNN model for the scene representation.

\subsection{Combined Model}
The Bayesian object relation matrix of the BORM is with the size of 150x150, which is first stretched to a vector of size 22500x1. The vector of BORM will be fed into the two fully connected layers and an discriminative object feature  $F_{BORM}$ with 512 dimensions (for Places365-7) or 2048 (For Places365-14) will be the feature representation of BORM.

Meanwhile, we combine the PlacesCNN model of the ResNet backbone with BORM, as the combined Bayesian object relation model (CBORM).

\subsection{Combined Classifier}
For the Places365-7 dataset, there are two streams. To fair compare with the object model \cite{pal2019deduce}, one stream is based on ResNet18 pretrained on the Places365 and finetuned on the Places365-7 dataset. The other stream is BORM. The feature representations of ResNet18 and BORM will be concatenated and fed into a two-layer FC network with a dimension of 512, and 7 respectively. 

Similarly, for the Places365-14 dataset, there are two streams. To fair compare with the word-embedding \cite{chen2019scene} model, while one stream is ResNet50, and the other is BORM.
The feature representations of ResNet50 and BORM will be fed into a two-layer FC network with a dimension of 512, and 14 respectively.

\section{Experimental Results}
\label{sec:exp}

\subsection{Experimental Settings}
We evaluated the proposed models on reduced SUN RGB-D and Places365 dataset. To be noticed, aim to investigate the generalizability of our model, we evaluate our model pretrained on Places365-7 dataset on the SUN RGB-D dataset without retraining it. We'll introduce the implementation details and training procedure and different experiment settings.

\subsubsection{Implementation Details}
For the PlacesCNN model, ResNet18 or ResNet50 architecture is adopted in our experiment for ablation study. The optimizer used is the Stochastic Gradient Descent (SGD) with an initial learning rate of 0.01, the momentum of 0.9, and the weight decay of 0.0001. We decrease the learning rate 10 times every 10 epoch, and every time when updating the learning rate, we reload the parameters which have the best accuracy before this timestamp. The total number of epoch during training is 40. To be noticed, we use the training sets of Places365-7 and Places365-14 dataset for learning the BORM statistically, while testing on the SUN RGB-D dataset, where the BORM is the same as the Places365-7 dataset.

\subsubsection{Dataset Settings}

\begin{table}[]
\centering
	\caption{Dataset split setting, where the number of training set and number of test set are listed below.}
	\label{tab:dataset_split}
	\begin{tabular}{l|l|l}
		\hline
		Dataset      & Training & Test  \\ \hline
		Places365-7  &  35000   & 701     \\ 
		Places365-14 &  75000   & 1500    \\ 
		SUN RGB-D    &  35000(from Places365-7)        & 2077    \\ \hline
	\end{tabular}
\end{table}
\ \\
\textbf{Places365 Dataset:}
In this paper, we use the reduced Places365 \cite{zhou2017places}  dataset to test our methods, since it is the most largest and challenging scene classification dataset yet, and it contains broad categories in the indoor environment. In the experiment, we only consider the indoor scene recognition. There are two different settings on the reduced Places365 dataset. The one is Places365 with 7 classes includes Corridor, Dinning Room, Kitchen, Living Room, Bedroom, Office, and Bathroom, denoted as Places365-7.  The test set setting is the same as the official dataset and described in \cite{pal2019deduce}.
In addition, we use the reduced Places365 with 14 indoor scenes in Home environment includes Wet bar, Home theater, Balcony, Closet, Kitchen, Bedroom, Playroom, Laundromat, Bathroom, Living Room, Home office, Dining room, Staircase, and Garage denoted as Places365-14. The dataset splitting follows the same setting as described in \cite{chen2019scene}. The dataset splition can be seen from Table. \ref{tab:dataset_split}.

\textbf{SUN RGB-D Dataset:}
SUN RGB-D dataset \cite{song2015sun} is a challenging dataset for scene understanding that contains not only RGB images but also depth information of each image. It contains 3784 images collected by Kinect V2 and 1159 collected by Intel RealSense. Moreover, it incorporates 1449 images from the NYUDepth V2 \cite{silberman2012indoor}, and 554 images from the Berkeley B3DO Dataset \cite{janoch2013category}, both captured by Kinect V1. Finally, it takes 3389 manually selected distinguished frames without significant motion blur from the SUN3D videos \cite{xiao2013sun3d} captured by Asus Xtion.

In our experiment, we mainly consider the indoor environment understanding. Therefore, we use the reduced SUN RGB-D dataset includes Office, Kitchen, Bedroom, Corridor, Bathroom, Living room, and Dining room, where the test set split is the same as the official dataset. There are 3741 RGB images in total for testing. And we test our model pretrained on the Place365-7 on SUN RGB-D dataset without retraining.

\subsection{Experimental Results}
\subsubsection{Effect of Object Knowledge}
\begin{figure}[tbp]
	\centering
	\includegraphics[width=0.4\textwidth]{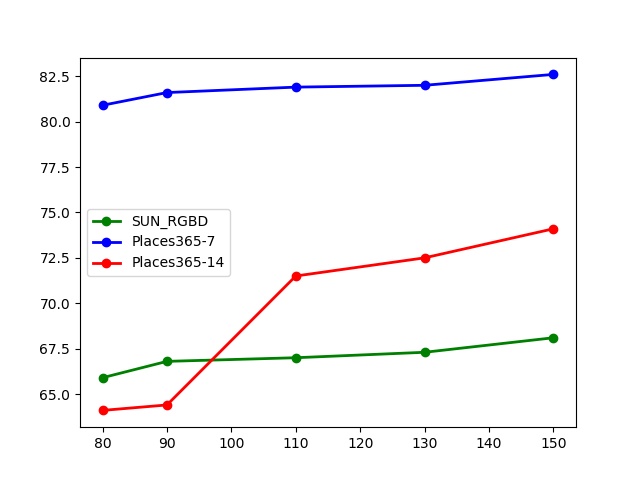}
	\caption{Ablation study of improved object model (IOM) with different number of object knowledge ranging from 80 to 150 categories as shown in horizontal axis. The vertical axis shows the accuracy on percentage.}
	\label{fig:om_ablation}
\end{figure}

As illustrated in Fig. \ref{fig:om_ablation}, a group of ablation studies have been conducted for evaluating the effect of object knowledge to indoor scene recognition on the Places365-14, Places365-7, and SUN RGB-D dataset, respectively. The x-axis represent the number of object information IOM have about the indoor scene and is added by 20 from 90 to 150 sequentially selected from the vector. Plus, IOM-80 is the baseline accuracy. Obviously, as the number of object information increases, the much better scene recognition accuracy is achieved, e.g., on the Places365-14 dataset, the IOM-150 reaches 74.1\% accuracy, which is \textbf{10.0\%} higher than IOM-80. This improvement shows the number of object information is proportional to scene recognition accuracy. Moreover, the comparison experiments between the IOM and OM on three datasets, shows an average of \textbf{20.5\%} improvements can be achieved. After analyzing the object categories of OM pretrained on the MS COCO, we observed there are only half of the object categories are related to indoor scenes. In contrast, the other half is related to outdoor scenes. Therefore, the relevance of object categories of object model with the scenes is essential for scene recognition, e.g., the information of elephant and giraffe in OM will not be valuable for indoor scene recognition. Similarly, the bus and train are not beneficial to indoor scene recognition.

\subsubsection{{Analysis of BORM}}

We conduct experiments for BORM and IOM in the reduced Places365-7 dataset, and results are displayed in Table \ref{tab:place365_7_sota}. The experiment results show the BORM and IOM model has an advantage over the OM and yields an average accuracy of 83.1\% and 82.4\%, surpassing the OM model about \textbf{20\%} accuracy. The experiment result proves that with more object knowledge about the surrounding environment, the greater scene recognition accuracy can be reached.  Then, we test the model pretrained on the Places365-7 dataset on the SUN RGB-D test set, and similar conclusion can be drawn. Moreover, the BORM outperforms the IOM with 0.7\% and 1.1\% accuracy on the Places365-7 and SUN RGB-D dataset, respectively, which validates the knowledge of the co-occurrences between object pairs and their probabilistic relation forms an important indoor scene representation.

Similarly, in the Table \ref{tab:place365_14_sota}, we have conducted experiments on the reduced Places365-14 dataset, and experiment results show the IOM and BORM tremendously improves the performance over OM with \textbf{27\%} accuracy. The results suggest the effectiveness of BORM and IOM over the OM especially when the number of scene classes of dataset is large.

\subsubsection{Performance Comparison}

\begin{table}[htbp]
	\centering
	\scriptsize
	\begin{minipage}[t]{.44\textwidth}
		\centering
		\caption{Comparison with the state-of-the-art methods on the reduced Places365-7 Dataset and SUN dataset of scene recognition accuracy}
		\label{tab:place365_7_sota}
		\begin{tabular}{c|ccc}
			\hline
			Method     & Config   & Acc(Places365-7) & Acc(SUN) \\ \hline
			\multirow{2}{*}{PlacesCNN \cite{zhou2017places}}     & ResNet18 & 80.4  & 63.3  \\
			& ResNet50 & 82.7 & 67.2   \\ \hline
			\multirow{3}{*}{Deduce \cite{pal2019deduce}}     &  $\Phi_{obj}$ (OM)      & 62.6 & 53.6      \\
			& $\Phi_{scene}$    & 87.3 & 66.8    \\
			& $\Phi_{comb.}$ & 88.1 & 70.1    \\ \hline
			\multirow{3}{*}{Ours} 	
			& IOM  & 82.4 & 68.1   \\
			& BORM & 83.1 & 69.2   \\ \cdashline{2-4}
			& CBORM& \textbf{90.1} & \textbf{72.1\%}   \\	\hline
		\end{tabular}
	\end{minipage}

	\hspace{1cm}

	\begin{minipage}[t]{.44\textwidth}
		\centering
		\caption{Comparison with the state-of-the-art methods on the reduced Places365-14 Dataset of scene recognition accuracy, the * indicates the re-implement of the method. }
		\label{tab:place365_14_sota}
		\begin{tabular}{c|cc}
			\hline
			Method                    & Config           & Acc \\ \hline
			\multirow{2}{*}{PlacesCNN \cite{zhou2017places}}     & ResNet18 & 76.0  \\
			& ResNet50 & 80.0   \\ \hline
			\multirow{1}{*}{Word2Vec \cite{chen2019scene}} 
			& ResNet50+Word2Vec         & 83.7    \\ \hline
            \multirow{1}{*}{{*}Deduce \cite{pal2019deduce}} 
            
            & $\Phi_{obj}$ (OM) 		   & 47.0	 \\ \hline
			\multirow{3}{*}{Ours}  
			& IOM          & 74.1    \\
			& BORM         & 74.9    \\ \cdashline{2-3}
			& CBORM & \textbf{85.8}				\\	\hline
		\end{tabular}
	\end{minipage}
\end{table}

As shown in Table \ref{tab:place365_7_sota}, we conduct the ablation study of using only the BORM model and the CBORM model. We found the CBORM yield an average accuracy of \textbf{90.1\%}, which greatly outperforms the ResNet18 and ResNet50 of PlacesCNN baselines about \textbf{10\%} and \textbf{8\%} respectively. Meanwhile, CBORM outperforms the BORM with 7\% accuracy and 3\% on the Places365-7 and SUN RGB-D dataset, respectively.


In comparison with the state of the art, Table \ref{tab:place365_7_sota} shows that the CBORM improves the scene recognition by \textbf{2.0\%} in the reduce Places365-7 dataset.  Also, as shown in Table \ref{tab:place365_14_sota}, the CBORM improve the recognition accuracy by \textbf{2.1\%} in the reduced Places365-14 dataset. Both results show the combined model achieves comparable results to some recent approaches that use the word-embedding method to extract the semantic meaning of the environment, or use the combination of scene and object representations for better scene understanding. Moreover, Table \ref{tab:place365_7_sota} shows the performance of our method over the method in \cite{pal2019deduce} with \textbf{2\%}, showing the excellent generalization ability of CBORM over other methods on the Reduced SUN RGB-D dataset.

These results demonstrate that CBORM is successful in recognizing the scene images with a competitive accuracy. This improved effectiveness of CBORM over the state-of-the-art justifies our reasonable assumption that relation of object pairs is an essential complementary information for indoor scene recognition.

\section{Conclusion and Future Work}
\label{sec:conclusion}
In this paper, we aim to transfer object knowledge from humans to indoor scene recognition. First, we propose the IOM that with rich and relevant object categories for indoor scene representation. Besides, inspired by the nature of scene-common and scene-specific object pairs, we establish a BORM that obtains the probabilistic relations among object pairs given various scene, where the probability of scene-specific object pairs will be enhanced and the probability of scene-common object pairs will be suppressed.  
Moreover, we utilize the PlacesCNN model with ResNet as a backbone network for classification, which demonstrates a substantial result on the scene understanding. Hence, we combine the PlacesCNN and proposed BORM as CBORM for a more interpretable scene recognition algorithm, and experiment results show our proposed method significantly outperforms the state-of-the-art methods. 

In the future, we plan to integrate our algorithm to real robots like mobile robots, and flying robots for the construction of the semantic map includes the label of the scene and detailed semantic information of the environment. The construction of a semantic map using the proposed scene recognition algorithm would be useful for navigation in unknown places for robots and humans because both the semantic label of the region and semantic meaning of the environment are provided. Furthermore, our system could be applied to autonomous robots and enabling them to assist humans in safety and rescue missions inside a house or a building.

\bibliographystyle{IEEEtran}

\bibliography{Ref}

\end{document}